# Transformers perform adaptive partial pooling

Vsevolod Kapatsinski
*University of Oregon*

## Abstract

Any language model must decide what to say in novel contexts based on information from similar contexts. But what about contexts that are not novel but merely infrequent? In hierarchical regression, the model's predictions for behavior in a context are affected by observations from similar contexts to the extent that 1) the current context is infrequent and 2) different contexts behave similarly. This is called adaptive partial pooling. This paper shows that next-word predictions of a transformer (GPT2) are affected by observations from outside the current context, but this pooling reduces with more training. Pooling is affected by context frequency, context number (type frequency) and context variability in a qualitatively similar way to hierarchical regression. However, there is a "sweet spot" in training at which the transformer best matches the behavior of hierarchical regression. This is the point at which the effect of context frequency on pooling is at maximum.

## Introduction

Language is creative. Therefore, any language learner, human or machine, needs to be able to deal with novel sentences. Furthermore, because the frequency distributions of linguistic units are Zipfian (Zipf, 1949), with a long tail of low-frequency types, most linguistic units will occur only once even in a very large corpus. The main task required of a language model, and the one it is primarily trained on, is to predict the next word given the preceding context. Yet, the Zipfian frequency distribution ensures that most contexts will only occur in the corpus once, and therefore will not provide enough observations to reliably estimate the probability distribution over upcoming words.

The rational solution to this conundrum is **adaptive partial pooling** (Gelman & Hill, 2007). Because observations of words in the current context are more relevant to inferring the probability distribution in this context that observations in other context, an ideal learner should prioritize such observations. However, to the extent that they are too few to perform reliable inference, the ideal learner should bring in observations from similar contexts, partially pooling evidence across contexts to the extent that 1) the current context has too few observations for reliable inference, and 2) contexts are informative about each others' next-word probability distributions. The need for adaptive partial pooling has also been recognized in computational linguistics since at least the ngram models of the 1980s, which used a long context to predict the next word but either 'backed off' to shorter contexts when the frequency of the longer context was too low (Katz, 1987) or, more appropriately, weighed the longer and shorter context's predictions based on the difference in frequency (Jelinek & Mercer, 1980).

In theoretical linguistics, several researchers have also argued recently that hierarchical regression provides a good model of the mental grammar underlying language production, specifically because it uses adaptive partial pooling (Breiss et al., 2025; Kapatsinski, 2021; Morgan & Levy, 2015; Zymet, 2019).

However, there are some reasons to doubt that adaptive partial pooling plays a role in large language models (LLMs) and the transformer architecture that underlies them. First, there may be less need for pooling. A large language model is trained on a much larger corpus than experienced by any human. Therefore, even rare contexts are experienced multiple times, potentially reducing the motivation to base estimates of the word probability distributions in such contexts on other contexts. Second, modern deep learning models have been argued to be large enough to 'memorize' (or, more properly, faithfully reconstruct) the whole corpus. For example, they show equal facility at learning a random labeling of an image dataset and a structured one, even though the latter affords better opportunities for generalization through data compression (Zhang et al., 2016). It has been argued that the` ability to memorize the whole corpus is essential for high prediction accuracy (Belkin et al., 2019) and especially so for accurate performance in rare contexts (Feldman, 2020), for both image classification and language learning.

Empirically, Houghton et al. (2025) show that large language models tend to match word order probabilities for specific binomials like *bread and butter* or *bishops* and *seamstresses* and aren't significantly sensitive to more abstract factors that influence word order probabilities across contexts. However, this may not be due to how these models do inference. Instead, at least some of these factors (e.g., word length or social status) are likely not to be represented by the models and so are not accessible to them as predictors of next-token probabilities.

There are also reasons to think that transformers do implement adaptive partial pooling. Transformers are like older connectionist models in that they update hidden-layer representations via error-driven / discriminative learning. This is a gradual process of differentiating hidden-layer representations for contexts that are associated with distinct output probability distributions in training. All contexts are initially represented similarly, and become differentiated to the extent needed to produce distinct predictions (Rogers & McClelland, 2004, 2008). In this process, broad classes are discriminated before smaller subclasses within them, in a way that is reminiscent of human semantic development (Rogers & McClelland, 2004, 2008) For example, animals are differentiated from plants before fish are discriminated from birds, and animals or plants with similar affordances (e.g., plants with no known uses) may never be discriminated at all.

The increasing differentiation of contexts over time suggests that the amount of pooling should decrease with training, such that information is pooled across increasingly narrow classes of contexts. With enough epochs of training, and enough representational capacity, probability distributions may be estimated based entirely on the current rare context. However, until this happens, the model should be performing something similar to partial pooling.

How can we detect partial pooling in a transformer? In statistics, adaptive partial pooling is implemented by hierarchical regression models. As described in Gelman & Hill (2007) for hierarchical linear regression, the partial pooling estimate for some characteristic (α) of a word (w) in a context (cx) would be estimated by Equation (1), which defines adaptive partial pooling mathematically:

$$(1)\quad \hat{\alpha}_{cx.w} = \frac{\frac{n_{cx.w}}{\sigma^2_{cx.w}}}{\frac{n_{cx.w}}{\sigma^2_{cx.w}}+\frac{1}{\sigma^2_{w}}}\alpha_{cx.w} + \frac{\frac{1}{\sigma^2_{w}}}{\frac{n_{cx.w}}{\sigma^2_{cx.w}}+\frac{1}{\sigma^2_{w}}}\alpha_{w}$$

Here, $n_{cx.w}$ is the number of observations (frequency) of the word *w* in context *cx*, $\sigma^2_{cx.w}$ is the observed variance of the characteristic (α) across observations of the word *w* in context *cx*, and $\sigma^2_{w}$ is the variance of average $\alpha_w$ across contexts. Equation (1) says that the estimate based on the current context ($\alpha_{cx.w}$) is weighted more when the context is frequent ($n_{cx.w}$ is high). Conversely, the estimate based on other contexts is weighted more when words tend not to differ in (α) across contexts.

If $\alpha_{cx.w}$ is probability of the word given the context, $\sigma^2_{cx.w}$ is irrelevant and therefore $n_{cx.w}$ and $\sigma^2_{w}$ are the only two predictors that matter for how much pooling across contexts a regression model would perform in estimating the probability. If they matter in the same way for probability estimates derived from a transformer, we have evidence that transformers too perform adaptive partial pooling.

The present paper therefore investigates whether large language models (and GPT-2 in particular) perform adaptive partial pooling, and specifically

- how the amount of pooling changes across epochs of training, and
- whether the amount of pooling in these models is sensitive to the same predictors as hierarchical regression.

Note that this is entirely a behavioral comparison, which I believe is the proper level for evaluating this hypothesis. Models (and organisms) can implement adaptive partial pooling in many different ways. What matters for performance (and survival) is that they do.

## Approach and Methods

Since the main interest of this paper is to compare the inferential mechanisms of a large language model to those of a hierarchical regression, we compare them on an artificial language that would be represented very similarly by the two model types.

Specifically, we focus on an artificial language in which next-word probability distributions are generated by a hierarchical regression model using two nested predictors. This ensures that the grammar of the language can be represented perfectly by a hierarchical regression model. In addition, the predictors influencing next-word probability distributions are tokens of the transformer. Thus, the large language model, like the regression model, will be treating them as basic units of the system, though (unlike regression) it can transform them internally.

Importantly, the tokens defining contexts are new (added to the tokenizer of GPT2), ensuring that novel contexts are indeed novel for the large language model despite its pretraining. This ensures that pretraining cannot generalize to the artificial language in question, which we confirm by the fact that the model has no expectation that the contexts will behave differently before training on the artificial language.

The language is also small enough for GPT2 to memorize, making it likely that the results would hold for larger models that have bigger representational capacity.

Finally, as in natural languages (Zipf, 1949), the distribution of context frequencies was Zipfian, following a power law with an exponent of 1.

Specifically, the language is defined by the following grammar, where {} are mutually-exclusive sets of tokens:

(2) $\{token_{1...10}\}$ X {A;B}
$\{token_{11...110}\}$ Y {A;B}

That is, the string always ended in A or B, preceded by X or Y ("penult"). One of the penults (X or Y) could be preceded by 10 distinct tokens, while the other could be preceded by 100 distinct tokens. None of the tokens preceding X could also precede Y and vice versa, i.e., contexts are nested within the X and Y classes. The tokens were added to the GPT-2 tokenizer to ensure that they were novel. For each replication, 1000 string instances were generated.

The number of A and B tokens was generated by sampling from a binomial distribution with context frequency as the number of trials and probability of A defined as in (3). That is, log odds of A started out at 0 and increased by *b* after Y and decreased by *b* after X (or vice versa for the other half of the runs). *b* was set to 1. The tokens preceding X or Y also changed the log odds of A by an amount drawn from a normal distribution with mean 0 and standard deviation *s*, which varied between 1 and 2.

$$(3)\quad p(A|cx) \propto logit^{-1}(b \times Penult + N(0, s_{token})))$$

Figure 1 shows the Zipfian context frequency distributions, for both the type-frequent context group (here, X) and the one of lower type frequency (here, Y). Equalizing token frequency across the two contexts requires higher token/type ratios for the Y group. The type frequency manipulation is motivated by findings that high type frequency makes a linguistic pattern more productive – more likely to be extended to new contexts by humans (e.g., Hayes et al., 2009; Jarosz et al., 2025; Olejarczuk & Kapatsinski, 2018). Surprisingly, however, Eq. 1 assigns no role for the number of distinct contexts.

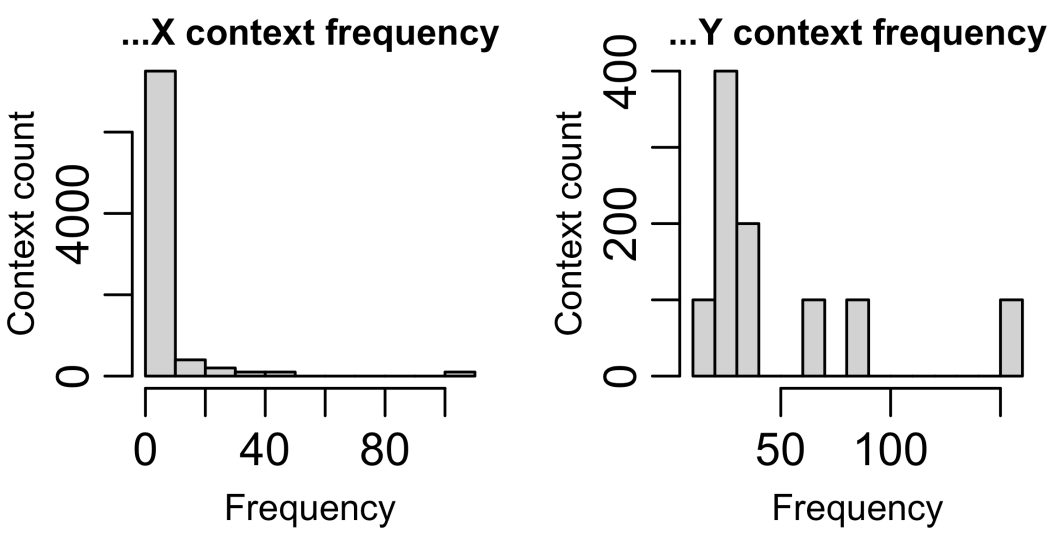

Figure 1: Context frequency distribution (one replication)

Figure 2 shows the distributions of next-word probabilities of A and B given each context. On the left, we see true probabilities that feed into the binomial sampling process, and on the right we see the observed probabilities resulting from sampling. Observed probabilities are more extreme than true ones.

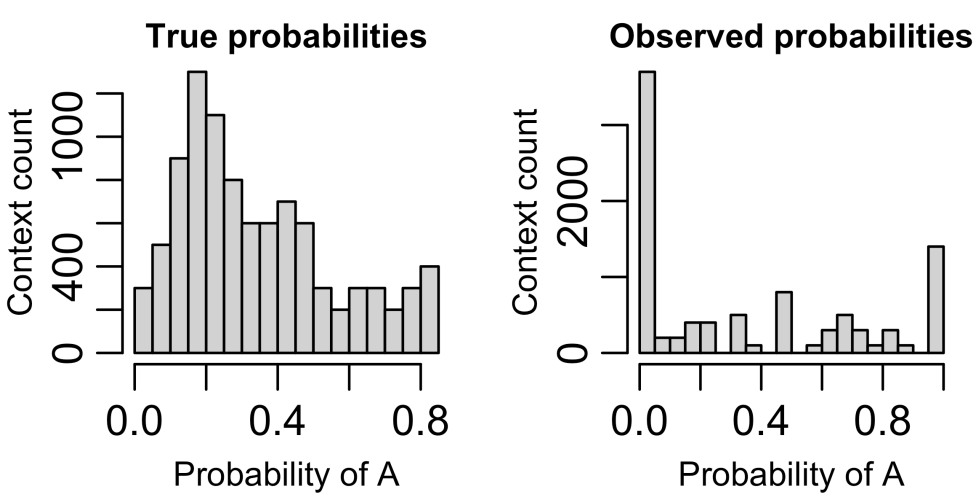

Figure 2: True and observed probabilities of A depending on the effect of group ($b = 1$; $s = 1$)

Models were run in Google colab using pytorch and GPT2 from transformers (version 4.37.2). The available version of GPT2 has been pretrained on the Internet (Radford et al., 2019). We train ("finetune") it on the artificial language. As noted above, the novel tokens 1-100 are added to the GPT2 tokenizer, ensuring that the contexts are novel to the model and so pretraining should not provide the model with relevant knowledge. Furthermore, the relations of penult and final are counterbalanced. The hierarchical logistic regression model had a random intercept for token and a fixed effect for penult (X/Y). It was a Bayesian model with normally distributed uninformative priors with $\sigma = 5$ on the fixed effects (intercept and penult) and a half-normal prior with $\sigma = 3$ on the normal distribution of random intercepts. Data analysis and graphing were performed in R. All code is available on OSF (https://osf.io/z42ea/overview?view_only=e2fa6d7889a041d6aa9c14e92b784315).

A dataset defined by *s* and frequency parameters was generated 50 times, with GPT and regression models fit to each dataset. GPT probability distributions (A vs. B) were generated and saved for each dataset at each epoch of training. We then investigate how model-inferred next-word probabilities differ across specific contexts depending on the observed probability distribution in that specific context, the probability distribution pooled across all contexts ending in the same word (X or Y), the model, *s*, the frequency of the specific context (token frequency), and the frequency characteristics of the group of contexts sharing the same final word (type frequency, token frequency, and token/type ratio). Both regression predictions and transformer predictions are probabilities, and therefore bounded between 0 and 1. We therefore used beta regression from the betareg package (Cribari Neto & Zeileis, 2010) for analyzing how predictions are affected by our independent variables. [1]

## Results

### Context infrequency increases pooling

Figure 3 (next page) shows that in both models, inferred probabilities diverge from observed context-specific probabilities (red line) in the direction of the probability for the group of contexts ending in the same penult (blue line). This divergence indicates pooling. The divergence is markedly greater in early GPT epochs, where there is more pooling than in regression. However, in both models, there is more pooling in rare contexts (freq <=10 here). This is especially obvious at the left and right edges, where the specific context probability is far from the group probability. During epoch 1, the model learns the effect of penult (final word of the context), but treats all contexts ending in the same word/token as the same. By epoch 20, the contexts ending in the same word have become differentiated.

Figure 4 shows that pooling decreases over the epochs. The effect of context frequency on pooling is in the same direction in GPT and hierarchical regression: there is, rationally, more reliance on out-of-context probabilities when the context is rare. However, the effect of context frequency on GPT predictions is initially larger than in regression (Figure 3, top) and eventually becomes smaller than in regression (Figure 3, middle row).

Figure 5 plots how the interactions of group and context-specific Observed probabilities with context frequency evolve across epochs of training, based on fitting the model in Table 1 (next page) to each epoch. The interactions grow in the expected direction – the effect of Group probability decreasing in frequent contexts, and the effect of context-specific probability increasing – until about Epoch 11 and then change course and diminish again.

[1] Refitting the full model (Tables 1-2) with linear regression on log odds leads to the same results. Linear regression does slightly better than beta regression on average, but the difference is very small (median $\Delta R^2 = 0.02\%$), and the predictions correlate almost perfectly ($r = 1$); see Additional Simulations in OSF.

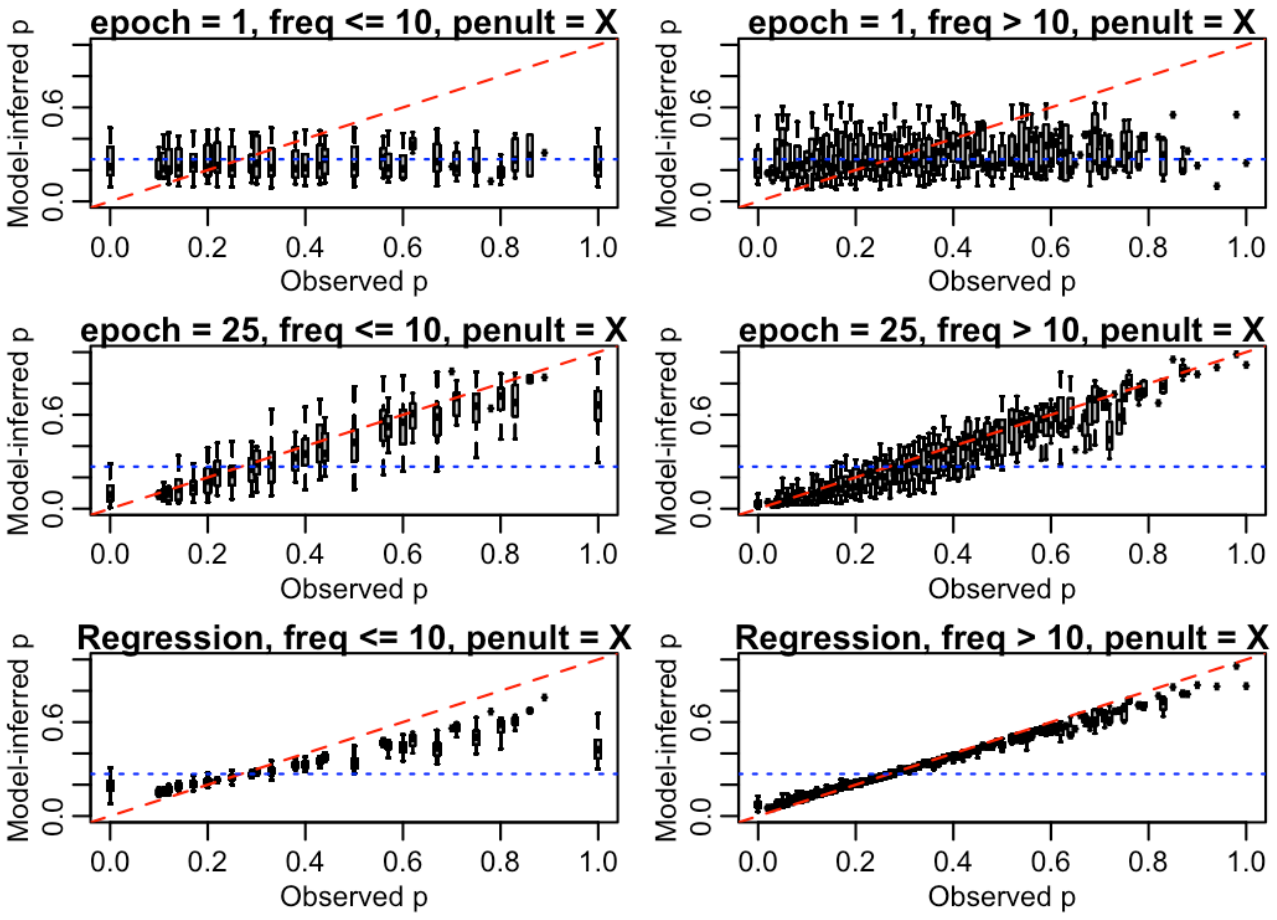

Figure 3: Change in inferred probability for contexts ending in X for the GPT-2 model (epoch = 1 vs. 20) and the corresponding inferred probabilities from hierarchical regression. Freq = Frequency of specific context. Red line: model probabilities = observed context-specific probabilities. Blue line: model probabilities = probability given the penult only (group probability).

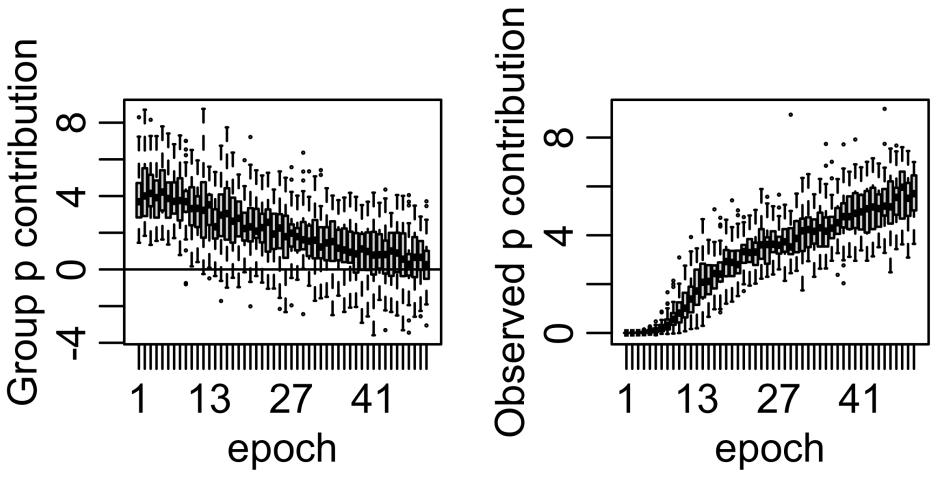

Figure 4: Left: Contribution of Group probability to GPT's Model-inferred probability across epochs of training (coefficient for Group p from beta regression). Horizontal line at zero. Right: Contribution of context-specific Observed probability across epochs.

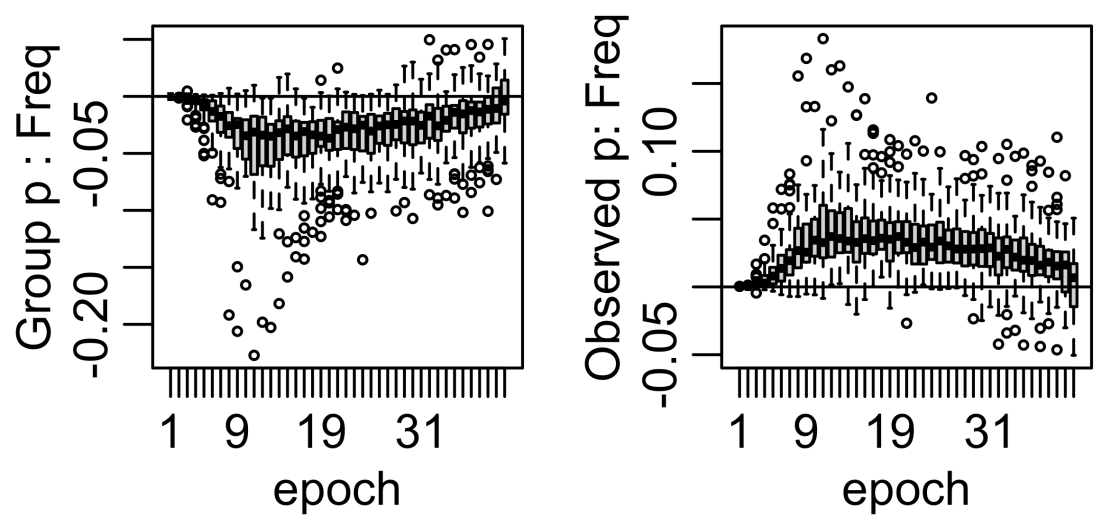

Figure 5: Changes in the interactions between context frequency and the context group probability (left) or context-specific Observed probability (right). Black line is at 0 = no interaction.

Since we know the probabilities that were used to generate the data, we can determine the point in training at which GPT fits them best. Figure 6 shows how the correlation between true and inferred probabilities changes over epochs of training. GPT achieves the best fit to true probabilities at epoch 11, around the point at which the interactions between context frequency and specific/pooled probabilities are the highest. Overall, GPT also tends to fit the probabilities worse than hierarchical regression.

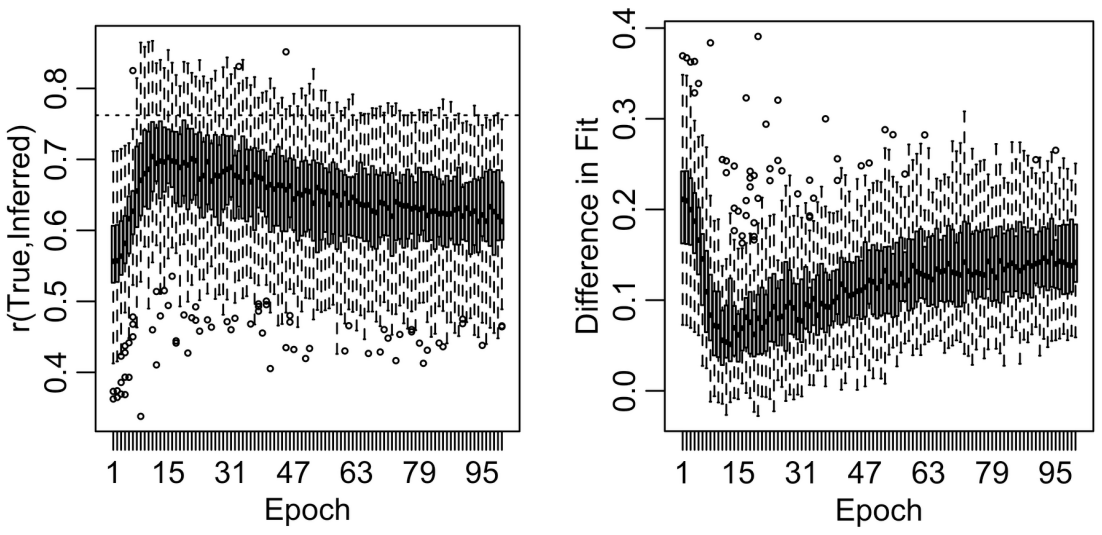

Figure 6: Correlations between model-inferred and true probabilities across epochs. Left: Boxes shows correlations with GPT-inferred probabilities. Dotted line shows the correlation with hierarchical-regression-inferred probabilities. Right: Difference in correlation with true probabilities between regression and GPT (fit to the same dataset). Differences above 0 = regression shows better fit.

Figure 4 showed how coefficients for context-specific and group probabilities evolve over the course of training. We saw that the contribution of group probabilities diminishes over time, but stays above 0, showing some partial pooling persisting into epoch 50. The interactions with context frequency also weaken with training in GPT, and dissipate if training continues, losing significance ($p > .05$) by Epoch 40.

How much training is 40-50 epochs? Figure 6 showed that fit to *true* probabilities declines after epoch 11. Therefore, models trained for more than 11 epochs can be considered fully trained (or overfitted) – they will never do any better at fitting the true probabilities. Thus, Figure 4 shows that fully-trained models do exhibit partial pooling, with a non-zero contribution of group probability.

Table 1 shows coefficients for Group probability, context-specific Observed probability and their interactions with frequency for GPT2 at epoch 11 and hierarchical regression. The results are qualitatively very similar. Thus, when fitting the true probabilities well, GPT performs adaptive partial pooling in a way very similar to hierarchical regression.

Table 1: Influences on inferred next-word probabilities from GPT2 (epoch 11) and hierarchical regression. All $p <$ .00001.

| | $b$(GPT$_{11}$) | $b$(Reg) |
|---|---:|---:|
| (Intercept) | -2.22 | -2.11 |
| Group p | 2.88 | 2.68 |
| Observed p | 1.41 | 1.34 |
| Group p : Freq | -0.06 | -0.035 |
| Observed p: Freq | 0.067 | 0.036 |

**Context diversity reduces pooling**

In both GPT and regression, when contexts within a group are more diverse in behavior (in regards to the next-word probability distributions), there is less pooling across the contexts: the effect of group probability on inferred probability is weaker (left panel of Figure 7), and the effect of context-specific probability is stronger (right panel of Figure 7). This effect of within-group between-context variance increases across epochs of training, until about epoch 15, with some decline thereafter (Figure 7). The dotted lines show the corresponding effects in hierarchical regression. Over the epochs, GPT approaches the regression effect.

Interactions between variance and the context-specific and group probabilities are shown in Table 2 for GPT (at epoch 18, which shows the best fit across variance with $b = 1$) and hierarchical regression. The interactions are very similar, and consistent with partial pooling occurring in both models.

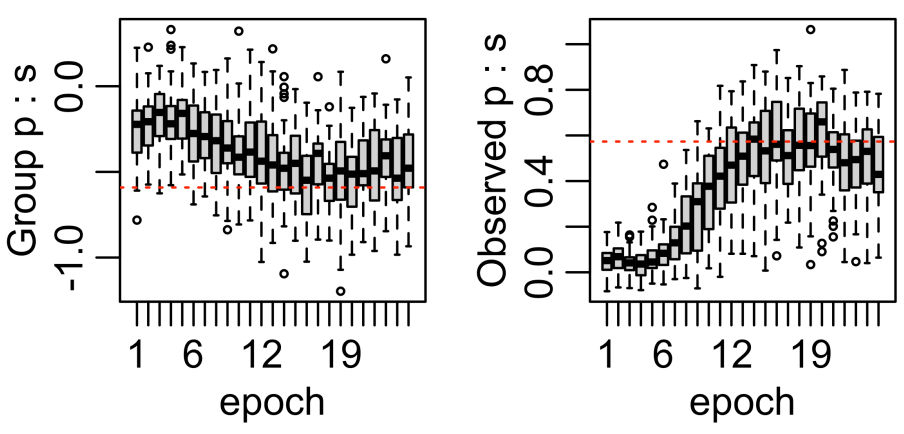

Figure 7: The influence of within-Group variance in next-word probability distributions on the effects of Group probability (left) and context-specific Observed probability (right) across epochs. Dotted red lines show the effect observed in hierarchical regression.

Table 2: The interactions between Group next-word probability, context-specific Observed probability and across-context variance in GPT and Hierarchical regression.

| | $b$(GPT$_{11}$) | $b$(Reg) |
|---|---|---|
| Group p : Var | -0.53 | -0.59 |
| Observed p: Var | 0.54 | 0.57 |

**Type frequency increases pooling and amplifies effects of token frequency**

Despite the lack of any terms for number of contexts in Eq (1), there is a strong effect of type frequency in hierarchical regression, as shown in Figure 8 (top left, red dotted line). The effect of a Group of contexts is stronger when the contexts in the group are more numerous, controlling for $s$, $b$, and relative token frequency of the groups. Conversely, the effect of specific-context Observed probability is weaker when the context belongs to a group of many contexts. These effects are also seen in GPT, but are weaker in that model.

The weakening of the effect of Group probability in frequent contexts (Figure 8, bottom left), and the corresponding strengthening of the effect of context-specific Observed probability (Figure 8, bottom right) also interacts with type frequency. Both effects of token frequency are stronger when type frequency is high in regression. In GPT, these interactions start at 0 but ramp up to the levels predicted by regression (bottom row of Figure 8).

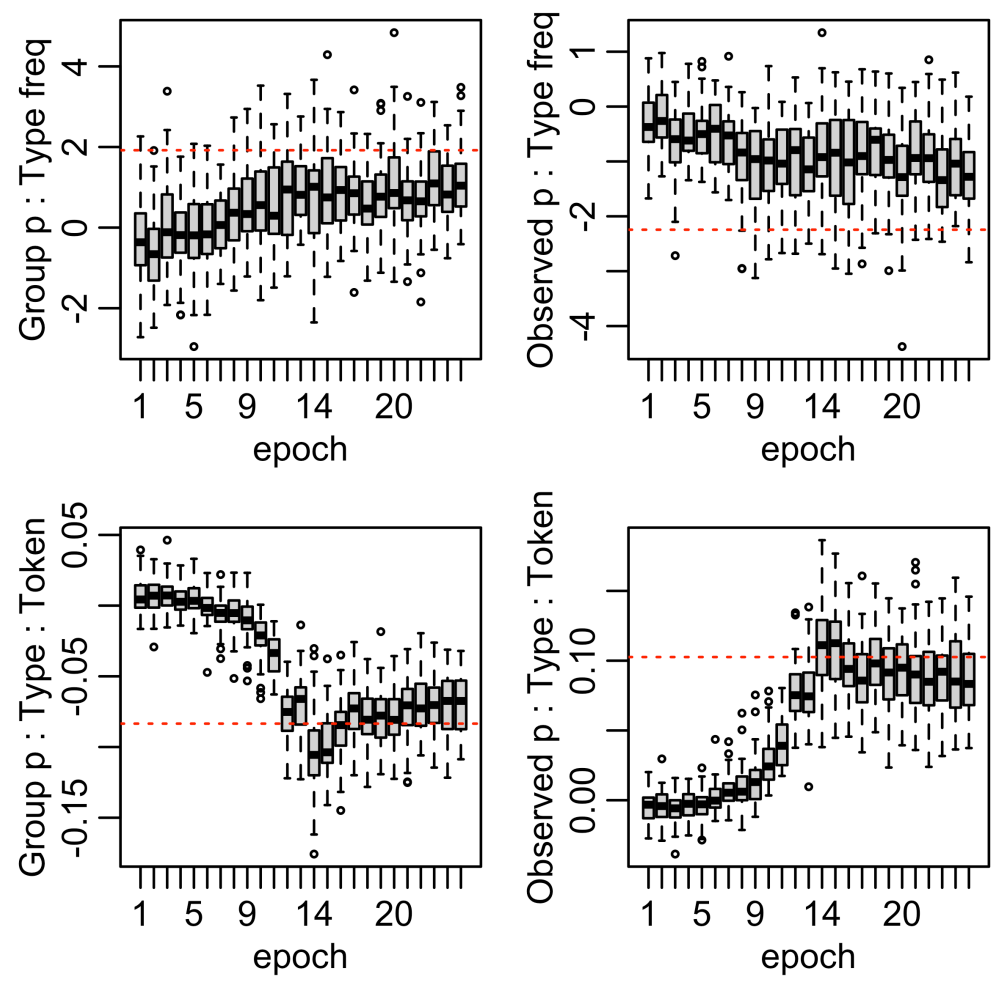

Figure 8: The effect of Group type frequency. Top row: Interactions with the effect of Group and Observed probabilities across epochs. Bottom row: Interactions with the interactions between context frequency and the Group and Observed probabilities.

## Discussion

In this paper, we demonstrated that adaptive partial pooling does emerge in transformers, and is similar to pooling in hierarchical regression. In both approaches to inference, evidence from the current context is (eventually) prioritized over evidence from other contexts. Figure 3 shows that, by epoch 25, GPT2 mostly uses the probabilities from the current context to infer next-word probabilities, although it starts out relying on the whole group of contexts ending in the same word. In both models, evidence from other contexts plays more of a role when the current context is rare (Figure 5). In both, more pooling is observed when contexts within a group tend to behave alike (Figure 7). In both, type frequency (the number of distinct contexts in a group) increases the amount of pooling (Figure 8).

This last effect has not been previously widely recognized for regression: Equation (1), taken from Gelman and Hill's (2007) influential textbook on hierarchical inference, makes no reference to the number of contexts within a group. However, it makes intuitive sense: the more similarly-behaving contexts in a group there are, the more evidence there is that new contexts in that group will also behave the same way. This is the same logic thought to underlie the effects of type frequency in human generalization: the greater the proportion of existing contexts that behave in a certain way, the greater the likelihood that a new context will behave in the same way (e.g., Kapatsinski, 2023).

Pooling in transformers diminishes over the epochs of training. The learning trajectory is consistent with the discriminative approach to learning theory, well supported in humans and animals (Hoppe et al., 2022). Adaptive partial pooling emerges in transformers because, like other error-driven connectionist models with hidden layers, they develop increasingly differentiated internal representations of input patterns (Rogers & McClelland, 2004). In this case, the model gradually learns which contexts behave differently with respect to next-word prediction (Figure 3). The learning dynamics look quite similar to the learning dynamics in the multilayer perceptron studied by Rogers and McClelland.

The design of the current experiment ensures some generality to this finding. First, the results are likely to hold for larger models because the current language is small enough for even GPT2 to represent comfortably. Second, they are also likely to hold for models with different pretraining. In the experiments above, contexts are defined by novel tokens that the model does not encounter in pretraining. Making X, Y, A and B novel tokens as well makes the model slower to learn that X and Y matter (this now takes 7-10 epochs; cf. Figure 9 vs. Figure 4). However, it does not change the overall dynamics: the model still differentiates penult-sharing contexts first, and only later differentiates contexts sharing a penult.

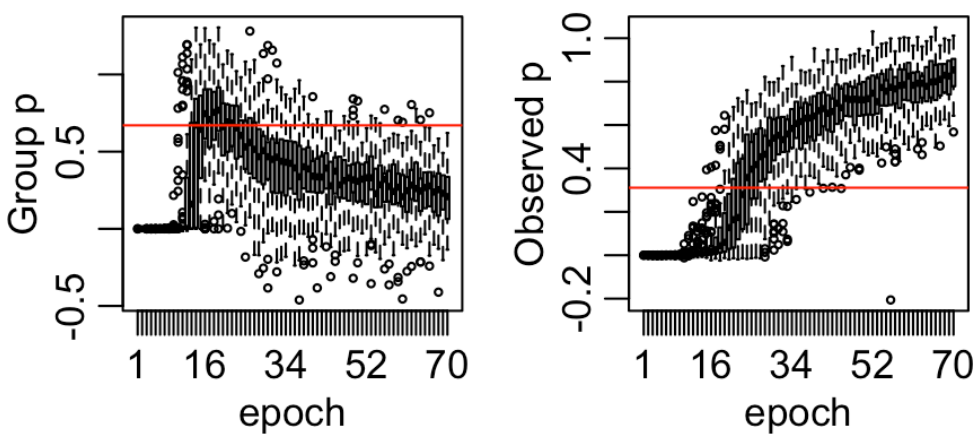

Figure 9: If penults are new, it takes a few epochs for the transformer to notice that they matter, but it still notices this first. Red horizontal lines show coefficients from regression.

Each epoch exposes the transformer to all of the relevant data whereas humans experience all of the data once. This makes frequency effects, robust in human processing, disappear over training (Heitmeier et al., 2024; Oh & Schuler, 2025). With respect to partial pooling, given enough training, a transformer will accumulate enough apparent evidence from even the rarest of contexts to trust that context's next-word probabilities. Modern large language models (LLMs) are not trained for multiple epochs, but they are trained on an extremely large corpus (Komatsuzaki, 2019; Touvron et al., 2023). Creating a larger corpus using the grammar in (2) would be exactly the same as training the model for multiple epochs on the current corpus because the grammar cares only about the categorical context properties *penult* and *antepenult*. Real grammars care about more than that (e.g., lexical semantics; Suttle and Goldberg, 2011). Nonetheless, grammars appear to be coarse-grained enough to be learned early in training an LLM. Hu et al. (2020) show that grammaticality does not decrease appreciably as training data decreases to 1 million words, which is orders of magnitude below current training corpus sizes. Recent work has shown that LLMs fit human reaction times worse as they grow larger and are trained more (Aoyama & Wilcox, 2025; Oh et al., 2022). Furthermore, Oh et al. (2024) show that this decrease in fit is concentrated in rare words and suggest that overtraining makes LLMs predict rare words with super-human accuracy. I suspect that this is also true of context frequency: with enough training, LLMs come to rely too much on rare specific contexts, compared to humans. From this perspective, there should be a sweet spot in LLM training that would make it most humanlike, and perhaps this spot is when the transformer algorithm makes best use of adaptive partial pooling to estimate the true next-word probabilities.

This sweet spot is the point at which the transformer shows strongest interactions between context frequency and pooling, i.e., the point at which the pooling is most adaptive. While it is not possible to know where this sweet spot is a priori for a given model and corpus, it may be possible to identify it by tracking the interaction of pooling with frequency. Importantly, pooling can be quantified without knowing true probabilities. One simply needs to measure deviation of the inferred next-token probabilities from the objective next-token probability distribution conditioned on the current context in the corpus, in the direction of unconditioned next-token probabilities.[2]

An important caveat is that the "true probabilities" are only true if the data are generated using the process assumed by (hierarchical) logistic regression, i.e., sampling from a binomial distribution defined by the true probabilities. Although this is a long-held assumption in probabilistic models of mental grammar (Kapatsinski, 2021; Morgan & Levy, 2015; Paolillo, 2001; Sankoff & Labov, 1979; Zymet, 2019), it has not been extensively tested empirically. Perhaps, the excellent performance of LLMs tells us that language production implements a less noisy sampling process.

For many next-token choice points, the probability distributions are highly polarized – the majority of contexts categorically favor continuation A, while a minority categorically favor continuation B (e.g., the suffixes *-ness* vs. *-ity*; Morgan and Levy, 2015; Zuraw, 2016). This is not the case for true probabilities in our simulations (Figure 2, left), though it is the case for *observed* probabilities (Figure 2, right). With a polarized *true* probability distribution, it could be safe to assume that a context in which the speaker chose A once is a context in which they will always choose A. Since current LLMs use very long contexts, it is possible that the choice given context becomes much more deterministic. In preliminary simulations, we do see GPT perform better when the underlying true distributions are highly polarized (e.g., $b = 3$), approaching the level of regression (with ~93% vs. 96% accuracy). Nonetheless, overtraining still decreases fit to true probabilities. This suggests that the "sweet spot" is real.

[2] Transforming probabilities to log odds interestingly eliminates this sweet spot in terms of the strength of interactions with frequency – they do not diminish with training (see the simulations in 'Additional Simulations' on OSF). However, this is only if we include observed probabilities of 0 and 1, for which observed logits are arbitrary and inferred logits grow more extreme with training.

## References

Aoyama, T., & Wilcox, E. (2025). Language models grow less humanlike beyond phase transition. arXiv preprint arXiv:2502.18802. https://doi.org/10.48550/arXiv.2502.18802

Belkin, M., Hsu, D., Ma, S., & Mandal, S. (2019). Reconciling modern machine-learning practice and the classical bias–variance trade-off. *Proceedings of the National Academy of Sciences*, *116*(32), 15849-15854. https://doi.org/10.1073/pnas.1903070116

Breiss, C. M., Hayes, B. P., Sundara, M., & Johnson, M. E. (2025). Modeling how suffixes are learned in infancy. *Cognitive Science*, *49*(3), e70047. https://doi.org/10.1111/cogs.70047

Cribari-Neto, F., & Zeileis, A. (2010). Beta regression in R. *Journal of Statistical Software*, 34(2), 1-24. https://doi.org/10.18637/jss.v034.i02

Feldman, V. (2020). Does learning require memorization? a short tale about a long tail. In *Proceedings of the 52nd annual ACM SIGACT Symposium on Theory of Computing* (pp. 954-959).

Gelman, A., & Hill, J. (2007). *Data analysis using regression and multilevel/hierarchical models*. Cambridge University Press.

Hayes, B., Zuraw, K., Siptár, P., & Londe, Z. (2009). Natural and unnatural constraints in Hungarian vowel harmony. *Language*, *85*(4), 822-863. https://doi.org/10.1353/lan.0.0169

Heitmeier, M., Chuang, Y. Y., Axen, S. D., & Baayen, R. H. (2024). Frequency effects in linear discriminative learning. *Frontiers in Human Neuroscience*, *17*, 1242720. https://doi.org/10.3389/fnhum.2023.1242720

Hoppe, D. B., Hendriks, P., Ramscar, M., & van Rij, J. (2022). An exploration of error-driven learning in simple two-layer networks from a discriminative learning perspective. *Behavior Research Methods*, *54*(5), 2221-2251. https://doi.org/10.3758/s13428-021-01711-5

Houghton, Z. N., Sagae, K., & Morgan, E. (2025, July). The role of abstract representations and observed preferences in the ordering of binomials in large language models. In *Proceedings of the 63rd Annual Meeting of the Association for Computational Linguistics (Volume 2: Short Papers)* (pp. 695-702). https://doi.org/10.18653/v1/2025.acl-short.55

Hu, J., Gauthier, J., Qian, P., Wilcox, E., & Levy, R. (2020, July). A systematic assessment of syntactic generalization in neural language models. In *Proceedings of the 58th Annual Meeting of the Association for Computational Linguistics* (pp. 1725-1744).

Jarosz, G., Hughes, C., Lamont, A., Prickett, B., Baird, M., Kim, S., & Nelson, M. (2025). Type and token frequency jointly drive learning of morphology. *Journal of Memory and Language*, 104666. https://doi.org/10.1016/j.jml.2025.104666

Jelinek, F., & Mercer, R. L. (1980). Interpolated estimation of Markov source parameters from sparse data. In: Gelsema, E.S., Kanal, L.N. (Eds.), *Pattern recognition in practice* (pp. 381-397). North-Holland.

Kapatsinski, V. (2021). Hierarchical inference in sound change: Words, sounds, and frequency of use. *Frontiers in Psychology*, *12*, 652664. https://doi.org/10.3389/fpsyg.2021.652664

Kapatsinski, V. (2023). Understanding the roles of type and token frequency in usage-based linguistics. In M. Díaz-Campos & S. Balasch (Eds.), *The handbook of usage-based linguistics* (pp.91-106). Wiley. https://doi.org/10.1002/9781119839859.ch5

Katz, S. (1987). Estimation of probabilities from sparse data for the language model component of a speech recognizer. *IEEE Transactions in Acoustics and Speech Signal Processing*, 35, 400–401. https://doi.org/10.1109/TASSP.1987.1165125

Komatsuzaki, A. (2019). One epoch is all you need. arXiv preprint *arXiv:1906.06669*. https://doi.org/10.48550/arXiv.1906.06669

Morgan, E., & Levy, R. (2015). Modeling idiosyncratic preferences: How generative knowledge and expression frequency jointly determine language structure. *Proceedings of the Annual Meeting of the Cognitive Science Society*, *37*, 1649-1654.

Oh, B.-D., Clark, C., & Schuler, W. (2022). Comparison of structural parsers and neural language models as surprisal estimators. *Frontiers in Artificial Intelligence*, 5. https://doi.org/10.3389/frai.2022.777963

Oh, B.-D., & Schuler, W. (2025). Dissociable frequency effects attenuate as large language model surprisal predictors improve. *Journal of Memory and Language*, *143*, 104645. https://doi.org/10.1016/j.jml.2025.104645

Oh, B.-D., Yue, S., & Schuler, W. (2024). Frequency explains the inverse correlation of large language models' size, training data amount, and surprisal's fit to reading times. In *Proceedings of the 18th Conference of the European Chapter of the Association for Computational Linguistics*, *Vol. 1: Long papers* (pp.2644–2663). https://doi.org/10.18653/v1/2024.eacl-long.162

Olejarczuk, P., & Kapatsinski, V. (2018). The metrical parse is guided by gradient phonotactics. *Phonology*, *35*(3), 367-405. https://doi.org/10.1017/S0952675718000106

Paolillo, J. C. (2001). *Analyzing linguistic variation: Statistical models and methods*. Stanford, CA: CSLI.

Radford, A., Wu, J., Child, R., Luan, D., Amodei, D., & Sutskever, I. (2019). Language models are unsupervised multitask learners. *OpenAI blog*, *1*(8), 9.

Rogers, T. T., & McClelland, J. L. (2004). *Semantic cognition: A parallel distributed processing approach*. MIT Press.

Rogers, T. T., & McClelland, J. L. (2008). Précis of *Semantic Cognition: A parallel distributed processing approach*. *Behavioral & Brain Sciences*, 31(6), 689-749. https://doi.org/10.1017/S0140525X0800589X

Sankoff, D., & Labov, W. (1979). On the uses of variable rules. *Language in Society*, *8*(2-3), 189-222. https://doi.org/10.1017/S0047404500007430

Suttle, L., & Goldberg, A. E. (2011). The partial productivity of constructions as induction. *Linguistics*, 49, 1237-1269. https://doi.org/10.1515/ling.2011.035

Touvron, H., Lavril, T., Izacard, G., Martinet, X., Lachaux, M. A., Lacroix, T., ... & Lample, G. (2023). Llama: Open and efficient foundation language models. *arXiv preprint arXiv:2302.13971*. https://doi.org/10.48550/arXiv.2302.1397

Zhang, C., Bengio, S., Hardt, M., Recht, B., & Vinyals, O. (2016). Understanding deep learning requires rethinking generalization. *arXiv preprint arXiv:1611.03530*. https://doi.org/10.48550/arXiv.1611.03530

Zipf, G. (1949). *Human behavior and the Principle of Least Effort*. Addison-Wesley.

Zuraw, K. (2016). Polarized variation. *Catalan Journal of Linguistics*, *15*, 145-171.

Zymet, J. (2019). *Lexical propensities in phonology: Corpus and experimental evidence, grammar, and learning* [Doctoral dissertation, UCLA]. Proquest: https://www.proquest.com/dissertations-theses/lexical-propensities-phonology-corpus/